# Comfort-Centered Design of a Lightweight and Backdrivable Knee Exoskeleton

Junlin Wang, Xiao Li, Tzu-Hao Huang, Shuangyue Yu, Yanjun Li, Tianyao Chen, Alessandra Carriero, Mooyeon Oh-Park, and Hao Su*, *Member, IEEE*

*Abstract*— This paper presents design principles for comfort-centered wearable robots and their application in a lightweight and backdrivable knee exoskeleton. The mitigation of discomfort is treated as mechanical design and control issues and three solutions are proposed in this paper: 1) a new wearable structure optimizes the strap attachment configuration and suit layout to ameliorate excessive shear forces of conventional wearable structure design; 2) rolling knee joint and double-hinge mechanisms reduce the misalignment in the sagittal and frontal plane, without increasing the mechanical complexity and inertia, respectively; 3) a low impedance mechanical transmission reduces the reflected inertia and damping of the actuator to human, thus the exoskeleton is highly-backdrivable. Kinematic simulations demonstrate that misalignment between the robot joint and knee joint can be reduced by 74% at maximum knee flexion. In experiments, the exoskeleton in the unpowered mode exhibits 1.03 Nm root mean square (RMS) low resistive torque. The torque control experiments demonstrate 0.31 Nm RMS torque tracking error in three human subjects.

## I. INTRODUCTION

IN the last two decades, exoskeletons have been heralded as one type of promising assistive device for performance augmentation of healthy individuals [1]–[9] and medical rehabilitations of patients with disabilities [10]–[15]. Metabolic reduction has been considered the primary metric for device evaluation and its feasibility has been successfully demonstrated in walkers [16]–[19] post-stroke patients with paretic limbs [20], load carriers [21], and joggers [22]. From a design perspective, wearable robots are typically composed of actuators, transmissions, and wearable structure. Most exoskeletons with electric actuators are generally classified in terms of wearable structures as either rigid, or soft or flexible designs. Rigid exoskeletons (e.g. ReWalk or Ekso Bionics) rely on rigid materials to deliver torque in perpendicular to the musculoskeletal structure. Soft exosuits [23], [24] use cable transmission and textile-based wearable structures to deliver power from the actuator to the human through linear forces along the musculoskeletal structure. This innovation minimizes the joint misalignment issue with great metabolic reduction benefit [16]. However, it has limitations due to high-pressure concentrations [25] and the absence of weight-support functionality [26]. Flexible exoskeleton designs [26], deliver torque-type assistance (instead of linear force) with flexible structures by combining the advantages of rigid exoskeletons and soft exosuits.

The challenges of widespread adoption of this technology, however, arise from the manifestation (and need for resolution) of the discomfort due to excessive weight, or restricted range of motion, or high-pressure concentration; as well as the difficulty to develop a synergistic control that can mechanically assist human and physiologically adapt to human performance. Comfort and risk mitigation [27], [28] have been identified as two of the key features to allow individuals to safely and independently ambulate or use exoskeletons.

We propose to use shear force produced by the exoskeleton, joint misalignment, and actuator backdrivability as the quantitative measurement for comfort. Our contribution of this paper includes: 1) a structural analysis and design of a knee exoskeleton that ameliorates excessive shear forces; 2) a mechanism design that reduces joint misalignment and minimizes the distal weight; 3) a novel lightweight, compact, and highly-backdrivable actuation system. The overall weight of the exoskeleton prototype is 3.2 kg and its on-board battery can power walking assistance for 1 hour. Our exoskeleton design is intended to augment human capability by providing moderate levels of assistance at optimal timing of walking gait cycles as this methodology has been proved to be effective and efficient [25]. Normalized peak knee torques of 80 kg able-bodied individuals during walking and sit-to-stand are typically reported as 40Nm and 80Nm respectively. The knee exoskeleton in this paper aims to provide walking assistance. The peak output torque is 16 Nm, which is equivalent to 40% of peak biological knee moment of an 80 kg healthy individual

* indicates corresponding author. Junlin Wang, Xiao Li, Tzu-Hao Huang, Shuangyue Yu, Yanjun Li and Hao Su are with Lab of Biomechatronics and Intelligent Robotics (BIRO), Department of Mechanical Engineering, The City University of New York, City College, NY, 10023, US (E-mail: hao.su@ccny.cuny.edu). Junlin Wang and Xiao Li are with also with BrainCo Inc.

Tianyao Chen is with Department of Biomedical Engineering, Catholic University of America, Washington, DC, 20064, US.

Alessandra Carriero is with the Department of Biomedical Engineering, The City University of New York, City College, 10031, US.

Mooyeon Oh-Park is with the Burke Rehabilitation Hospital, NY, 10605, US.

Digital Object Identifier (DOI): see top of this page.

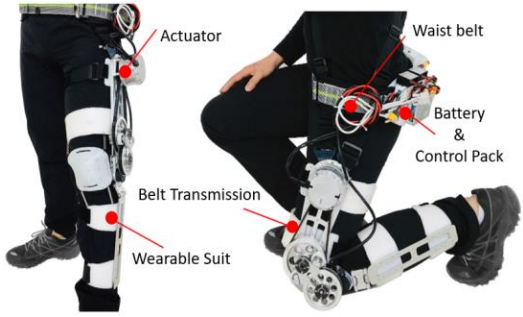

Fig. 1. A user wears the lightweight, compact and backdrivable knee exoskeleton in standing and kneeling postures, demonstrating its range of motion and compliance with human motion.

during walking [29], [30]. 16 Nm is also no less than the required torque to restore knee torque of paretic side knee of patients with stroke [31], [32].

## II. WEARABLE SUIT DESIGN

The wearable suit transmits the assistive torque generated by the actuator to the human body, and its design has a significant impact on the wearer's comfort. Our proposed design optimizes comfort in different parts of the suit. First, the attachment components are configured to distribute the assistive torque to the user by applying minimal pressure on the body. In addition, the direction of force is mostly perpendicular to the leg due to the mechanical frame design. Second, a rolling joint mechanism is implemented at the exoskeleton's knee joint so that the undesired forces caused by joint misalignment in the sagittal plane can be reduced without significantly increasing the mechanical complexity. Third, the joint misalignment in the frontal plane is mitigated by a double-hinge mechanism located on exoskeleton's calf frame. Finally, the attachment components are designed to precisely fit different individuals, so that the undesired impact load during exoskeleton's assistance is decreased. These features together provide a comfortable experience for the wearer. Detailed designs are illustrated in the following section.

### A. Suit Layout

The suit design is based on two considerations: the configuration of attachment components and the layout of the mechanical frame. The former factor mainly determines the way in which the device-generated joint torque is transmitted to the human body, and the latter affects the magnitude of certain undesirable loads.

The design of the strap attachment configuration is based on a force analysis of the human-exoskeleton model in the sagittal plane. Analysis results indicate that 4 attachment points (2 on the thigh and 2 on the calf) are preferred for minimizing the undesirable interaction force. Fig. 2 shows the analysis process as the model is reduced to a rigid body system of 4 segments. $F_p$ represents perpendicular interaction forces at attachment locations; $N$ stands for internal forces at joints; $\tau$ is the torque produced at the joints. The attachment component, typically a strap wrapped and tightened around the leg, applies both perpendicular and tangential forces on the limb. Such a condition leads to an underdetermined linear system with infinite solutions for the interaction forces. However, the attachment mechanism only allows small amount of tangential force in the form of friction, whose magnitude is negligible compared with the perpendicular force at the same location. Removing the tangential forces (Fig. 2 (c)) yields a consistent and overdetermined system with a unique solution. The solution shows that under the same loading condition, the magnitude of forces at contact points decrease as the distance between 2 contact points on the same body section increases. This indicates that with a 4-attachment layout, the proximal and distal attachment locations on the same body section should be as far away from each other as possible, so that the forces applied on human limb can be minimized given the same torque output. On the other hand, reducing the number of attachment points would make the system inconsistent, yielding no solution for the interaction forces. Having only one attachment on either thigh or calf would result in excessively high forces at the attachment location. This conclusion only applies to a knee exoskeleton with no additional rigid attachments to the wearer's torso or foot. Following the analysis, our configuration places the distal thigh attachment component and proximal calf attachment component at the closest distance to knee joint while securing knee flexion clearance; the proximal thigh attachment is placed near the groin and the distal calf attachment is placed just above the ankle joint. The exact value for the above locations on the mechanical frame are set based on 95th percentile body size data [33]. An improved lateral-support design shown in Fig. 3 (c) is proposed to mitigate the undesired twisting force in human-exoskeleton interaction.

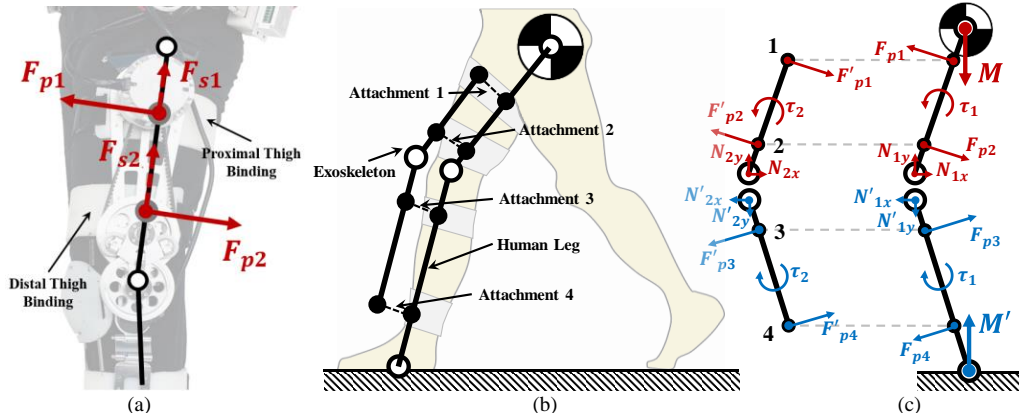

Fig. 2. Force analysis of human-exoskeleton interaction to optimize attachment configuration in terms of minimal undesired force. (a) The exoskeleton transmits assistive torque to the human leg as forces perpendicular to the leg ($F_p$) at attachment locations. Potential tangential forces ($F_s$) are caused by joint misalignment is ignored in attachment configuration design due to their limited magnitude. (b) The loading condition of the human-exoskeleton interaction can be simplified based on spring-loaded inverted pendulum model [34]. (c) Free body diagram of the rigid body system with the 4-attachment layout. Removing any one of the 4 attachments turns the system inconsistent with no solution, leading to excessive load at the other attachment locations.

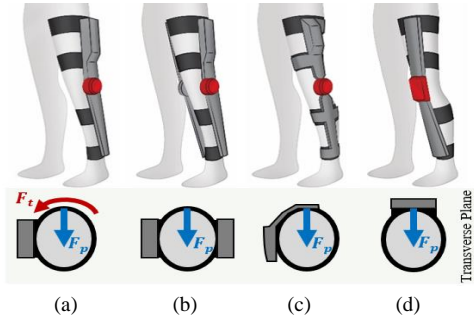

Fig. 3. Illustration of different knee exoskeleton layouts to demonstrate the advantages of the improved lateral-support design. (a) Lateral-support layout is most commonly used in multi-joint exoskeletons. The mechanical frame extends along the lateral side of the leg and transmits torque through soft attachment component (black) as $F_p$. The undesired twisting force $F_t$ is also generated as side-effect. (b) Two-side-support has an additional mechanical frame on medial side of leg, balancing the force transmission and avoiding undesired $F_t$. However, extra components on medial side causes interference during adduction movement. (c) Our improved lateral-support design has rigid attachment component extending from the frame to either anterior or posterior side of the leg. It avoids $F_t$ without introducing mechanical interference. (d) Anterior-support layout has the entire mechanical frame on the anterior side of leg. It generates the least undesired force compared to other layouts, but the implementation is limited by the complexity.

### B. Rolling Knee Joint

One major challenge in designing a comfortable exoskeleton is the misalignment between the human and the device joint. As illustrated in the previous section, when the joints are properly aligned, the lower limb and exoskeleton forms a 1-DOF multi-linkage system. As misalignment appears, the system becomes mechanically over constrained, causing an undesired tangential force at the attachment locations and excessive internal force at the human knee. Mitigating misalignment is challenging because the human knee joint has complex rotating mechanism combined rotation, rolling, and sliding [36], [37]. Many approaches have been explored to improve the alignment. One popular concept is to use an under-actuated mechanism that provides free-moving knee joint rotation center without affecting torque transmission [26], [38]–[42] ensuring alignment with the extra unactuated DOFs. Meanwhile, the designs without any rigid connection between the thigh frame and calf frame also effectively avoid misalignment. A different approach is approximating the human knee with 1-DOF mechanism to reduce misalignment [43], [44] without implementing underactuated mechanism [14], [45] Such designs typically utilize mechanisms like a 4-bar linkage or rolling cam(s), which provide a trajectory of rotation center similar to that of a human's.

In our design, we quantitatively analyzed the misalignment effect by means of simulation and accordingly proposed a rolling knee joint mechanism (Fig. 4 (a)) which can reduce misalignment when compared to a conventional revolute joint. Different from the design principles in [38], we emphasize on achieving a balance between misalignment mitigation and mechanism simplicity. The quantification is implemented in MATLAB/Simulink (Fig. 4 (b)). The kinematics of the exoskeleton-limb chain is modeled in Simscape Multibody. This section mainly focuses on the misalignment in the sagittal plane. Since knee motion in the frontal and transverse plane is relatively small, it is negligible in this model.

We adopt a classical human knee joint model that describes the relative motion of femoral condyle with respect to the tibial condyle as an ellipse rolling and sliding simultaneously along a flat surface. This model is first proposed in [46] by observing the magnetic resonance imaging images of 24 knees. The sliding ratio in this model, which is defined as the ratio between rolling distance and sliding distance can be adjusted to define a 1-DOF joint which can emulate the complex movements in the biological knee [47]. The two mechanical frames of the exoskeleton are modeled as rigid linkages fixed with the center of a roller on one side and the centers of two attachments on the other side. The couplings between the four attachments and thigh/shank are set as a prismatic joint because the sliding deformation is more obvious than the revolution deformation according to our observation in experiments. If the thigh is viewed as ground, the entire closed-loop kinematic chain possesses 9 rigid bodies (one thigh, one shank, two spurred gears which form the rolling joint, two mechanical frames and four attachments), one 1-DOF human knee joint, one 1-DOF rolling joint (exoskeleton knee joint), four 1-DOF prismatic joints between attachments and thigh or shank (two of them are passive constraints), and six 0-DOF fixed joint between mechanical frames and spurred gears or attachments. Therefore, the DOF of the system is one. More details about the simulation environment can be found in [48].

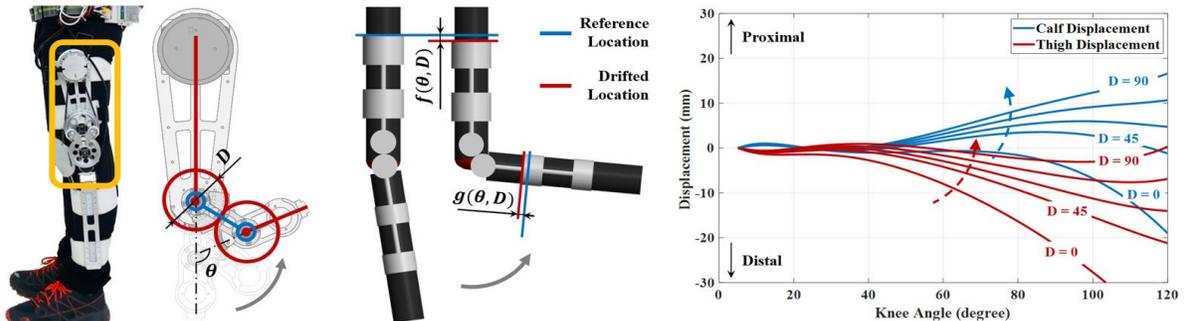

Fig. 4. Simulation of sagittal plane misalignment effect with rolling joint design to minimize undesired tangential force. (a) Rolling joint mechanism. The design functions as combined rolling and rotating, with both roller having the diameter $D$. (b) Simulation model with human leg colored in black and exoskeleton in grey. Misalignment between human joint and rolling joint causes the prismatic joints to slide along thigh model and calf model by the amount of $f(\theta, D)$ and $g(\theta, D)$ respectively. (c) Simulation results of frame drifting caused by misalignment under different knee angle $\theta$ and roller diameter $D$.

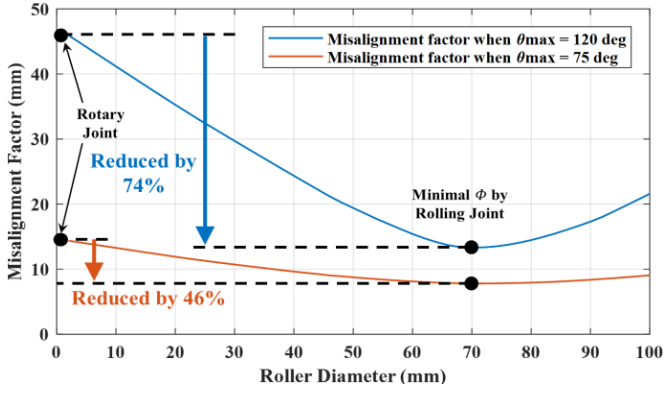

Fig. 5. The optimization of displacement factor $\Phi$ over different roller diameter $D$. The misalignment effect characterized by $\Phi$ is reduced by 74% and 46% when $\theta_{max}$ is 120° and 75°, respectively.

The displacement between frames and thigh or shank are the embodiment of the sliding deformation caused by misalignment. They may cause undesired interactive forces and discomfort during movement. We treat the roller diameter D as the design parameter to minimize the misalignment without employing other complex mechanisms. The conventional revolution joint can be regard as a special case of this model and its deformation performance in simulation is demonstrated in Fig. 4 (c) as the curves marked by "D=0". When the diameter of the rolling joint mechanism increases, the attachment deformation on thigh and calf both shift from distal drifting towards proximal drifting (shown in Fig. 4 (c) by curves marked with "D=45" and "D=90"). To evaluate the overall misalignment on both thigh and calf, the misalignment factor $\Phi$ is defined as follows:

$$\Phi(D) = \max_{5° \leq \theta \leq \theta_{max}} \sqrt{\left|f(\theta, x)\right|^2 + \left|g(\theta, x)\right|^2}\bigg|_{x=D} \quad (1)$$

where $\theta_{max}$ represent the maximum knee flexion angle considered, $f(\theta, x)$ and $g(\theta, x)$ respectively represent the amount of thigh and calf attachment, determined by knee angle $\theta$ and roller diameter D. Fig. 5 demonstrates that with the proper choice of D, introducing the rolling joint can reduce the misalignment factor $\Phi$ by approximately 74% when $\theta_{max}$=120° (around the maximum flexion angle of human) and 46% when $\theta_{max}$=75° (around the maximum knee flexion angle during normal walking) in the simulation.

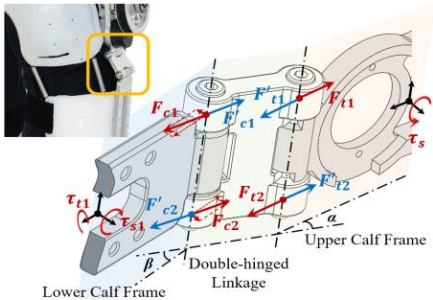

Fig. 6. Double-hinge connection on calf brace that mitigates frontal plane misalignment. Assistive torque is effectively transmitted through this underactuated mechanism between proximal and distal calf brace.

The roller diameter chosen in our design is 64 mm, with the rolling relation constrained by a pair of sectional spurred gear with the same diameter. Details of mechanical design is demonstrated in Fig. 9.

### C. Frontal Plane Misalignment Mitigation

In addition to the sagittal plane knee joint misalignment described in the previous section, the misalignment in the frontal plane also causes discomfort if not treated appropriately. During walking, the human knee was observed to have up to 10 degrees of varus and 4 degrees of valgus [49], while the exoskeleton's joint typically has no DOF in the frontal plane. We implemented a double-hinge mechanism on the mechanical frame between the rolling knee joint and the proximal calf attachment point, which provides 2 unactuated DOFs in the frontal plane to mitigate the misalignment (Fig. 6). The extra flexibility accommodates knee varus and valgus caused by both human knee rotations in the frontal plane, and the physical differences among subjects. Revolute dampers and mechanical hard stops are added to the hinge shafts to improve stability and preserve the potential anti-valgus function, respectively. When assistive torque $\tau_s$ in the sagittal plane is transmitted through the double-hinge mechanism from proximal calf frame to distal calf frame, the resultant torque applied on the calf is:

$$\begin{cases} \tau_{s1} = \tau_s \cdot \cos(\alpha - \beta) \\ \tau_{t1} = \tau_s \cdot \sin(\alpha - \beta) \end{cases} \quad (2)$$

where $\tau_{s1}$ and $\tau_{t1}$ are assistive torque about the instantaneous knee joint axis and undesired torque twisting the calf, respectively; $\alpha$ is the angle between the hinged linkage and proximal calf frame, and $\beta$ is the angle between the linkage and distal calf frame. It indicates that the assistive torque transmission efficiency through the mechanism only depends on the angle between proximal and distal frame, e.g. the varus and valgus angle of wearer's knee. Even with maximum varus, the mechanism still transmits 98.5% of the device-generated torque to user's calf about the knee rotation axis.

### D. Anthropomorphically-Customized Thermoplastic Brace

The design of attachment braces significantly influences the comfort and efficiency of the exoskeleton because it transmits device-generated torque to the human. If the rigid brace does not fit the shape of human lib, the gaps between the brace inner surface and wearer's leg create a backlash problem. In such a case, a pre-loading must be applied until the human's soft tissue is compressed to have proper contact with the brace. This pre-loading period creates a dead zone for controls during the switching between extension and flexion phase and introduces uncomfortable impact on the user. [14] implemented a high-level control scheme with pre-loaded torque profile to improve a similar problem.

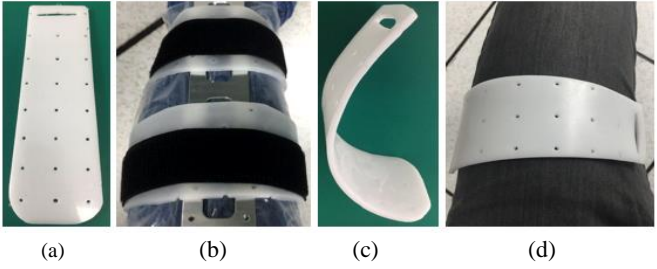

Fig. 7. Procedure of customizable braces as the mechanical interface attachment. (a) The unprocessed thermoplastic is machined to form raw part of universal shape. (b) Raw part is heated and applied on user's leg with the mechanical frame. (c) Raw part cools down and forms customized brace. (d) Precise fit with the user's limb.

In our design, we adopt thermoplastic (Fig. 7) as the material of attachment braces, the thermoplastic braces can precisely fit different individuals, reducing the control dead zone and undesired impact. In addition, a simplified tightening and buckling design are incorporated, reducing the donning/doffing time of the entire system to 2 minutes (Fig. 8).

## III. ACTUATION AND ELECTRONICS

Besides the wearable suit design, the actuation system also plays critical role for comfort-centered design. First, the backdrivability of the actuation subsystem determines the transparency of the exoskeleton when power is off. Second, excessive system weight and inertia about the human center of mass (CoM) costs extra energy cost for the wearer and obstruct human motion. Therefore, the design principle for actuation and electronics are high backdrivability, lightweight design, and low inertia.

### A. 2-Stage Timing Belt Transmission

We developed and integrated a 2-stage timing belt transmission with the rolling knee joint mechanism (Fig. 9). The ladder tooth synchronous belt is fabricated from Neoprene rubber and glass fiber rope that has a long lifecycle. The input axis of the 1st stage reducer (S1), aligned with the actuator, is located at the proximal end of thigh brace; the output pulley of S1 and the input pulley of 2nd stage reducer (S2) are concentric with the thigh roller center; the output pulley of S2 aligns with calf roller center. The S2 output pulley is fixed on calf frame and transmit the motor-generated torque to the frame. The output torque of the exoskeleton is defined on the instantaneous center of rotation of the calf roller in our paper. Assuming the radius of the input and output pulley is $r_1$ and $R_1$ of the 1st stage reducer, and $r_2$ and $R_2$ of the 2nd stage reducer, the effective lever arm of the belt transmission is $(R_2+r_2)/2$ (see Fig. 9). As a result, the total gear reduction ratio is

$$i_o = i_1(i_2+1)/2 \qquad (3)$$

where $i_1 = r_1/R_1$ and $i_2 = r_2/R_2$ are reduction ratio of S1 and S2 respectively. Considering the overall profile of the transmission and the reliability of involved transmission components, the transmission ratios of S1 and S2 are set at 4 and 3.43 respectively, leading to a total reduction ratio of 8.85. Paired with a highly integrated torque-dense actuator, the transmission generates a peak torque of 15.93 Nm and rated torque of 5.99 Nm. The distance between the S1 input and output shafts can be adjusted to tension the S1 timing belt. As the distance between input and output shafts of S2 is a fixed design parameter (roller diameter D), a pair of adjustable pulleys are used to tension the belt. The implementation of the high torque density actuator reduces the system weight by eliminating a conventional metal gear head (e.g. planetary gear or Harmonic Drive). In addition, it requires a low reduction ratio to reach high torque output, reducing the friction resistance within transmission and improving backdrivability.

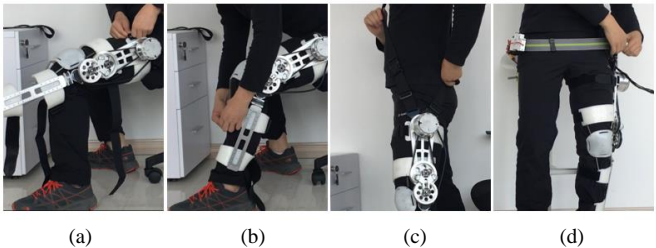

Fig. 8. Snapshots of the exoskeleton donning process that only takes less than 2 minutes without others' help. (a) - (b) First, the wearer binds exoskeleton at thigh and calf, which takes approximately 60 seconds. (c) An optional shoulder belt is then connected to prevent downward slip of the exoskeleton, and it takes an extra 18 seconds. (d) Finally, the wearer puts on the waist belt and connects power and signal cords with approximately 30 seconds.

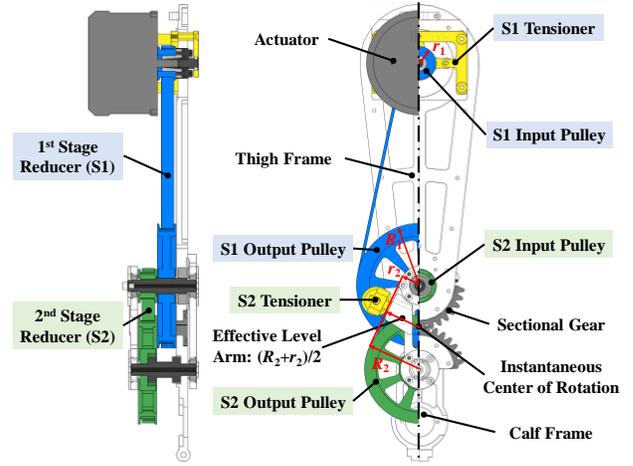

Fig. 9. Two-stage timing belt transmission system. This compact and simple system can amplify the torque of the electric motor to meet the assistive requirement. Moreover, through power transmission, the electric motor can be installed at the proximal end of thigh frame. The inertia of the exoskeleton can be significantly reduced by this configuration.

### B. Minimization of Distal Mass Distribution

It has been demonstrated by [50] that exoskeletons should have low inertia about the human's CoM. Therefore, on top of the lightweight design of each component, we position the majority of the system mass close to human CoM (Fig. 10). The 2-stage timing belt transmission design allows the actuator to be placed at the proximal end of the thigh frame. The battery and onboard processer are installed on the waist belt, which is approximately 50 mm below the human CoM. Compared to using a conventional design that integrates all components into the on-leg unit, the distributed-mass design reduces the inertia about the human COM in the sagittal plane by approximately 0.074 kg·m$^2$.

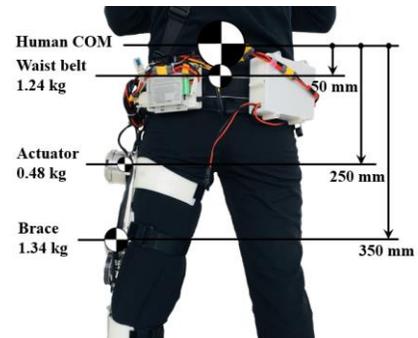

Fig. 10. The actuator is placed in the proximity to the center of mass to mitigate distal mass distribution to reduce metabolic cost burden.

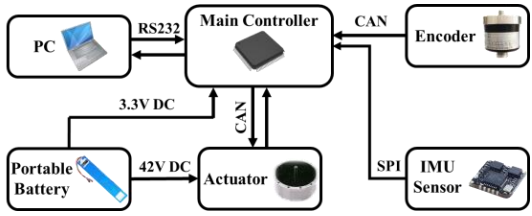

Fig. 11. The electronic hardware architecture of exoskeleton.

### C. Control Electronics and Communication

Fig. 11 illustrates the control electronics and communication for the real-time exoskeleton control. The torque control of the exoskeleton is achieved by the current motor controller implement in the motor module. The assistive control is based on the gait cycle detection through the IMU sensor (HI219M, HiPNUC, Inc.) and the data is processed in real time by RS-232. The communication between the main controller, motor module, and encoder are Controller Area Network (CAN) bus. The integrated drive electronics of the smart actuator aggregates the current, motor velocity, motor position.

The assistive control strategy is developed to follow 40% of the biological knee extension moment profile for stance phase gait assistance of hemiplegic stroke patients. The gyroscope of the IMU sensor measures the angular velocity of foot to detect the gait cycle. The control system diagram is shown in Fig. 12.

## IV. EXPERIMENT RESULTS

Extensive experiments were conducted to characterize and evaluate the performance of the exoskeleton in terms of its force tracking, mechanical transparency, assistive control strategy with human subjects.

### A. Torque Control Calibration

Due to the low reduction ratio of our transmission system,

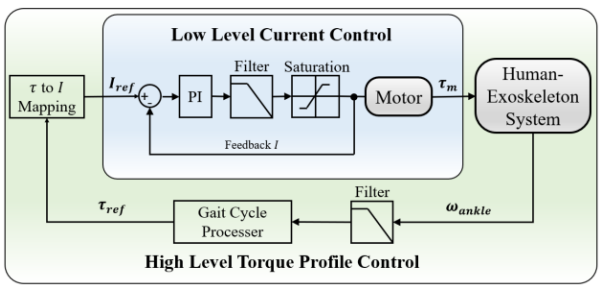

Fig. 12. Block diagram of assistive control strategy with high-level torque profile control and low-level current control.

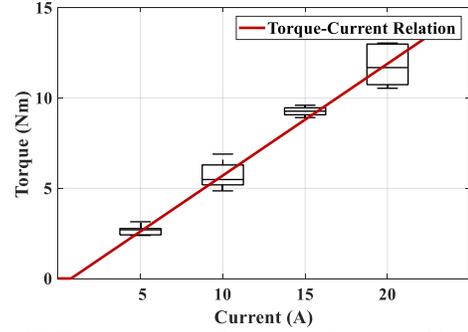

Fig. 13. The regression for the torque and current calibration.

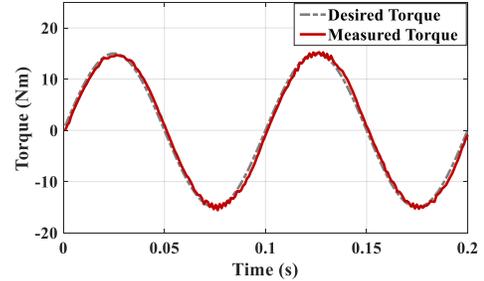

Fig. 14. Tracking performance of 10 Hz sine torque reference trajectory. The root mean square (RMS) error of torque tracking is 0.88 Nm, 2.93% of the ±15Nm tracking magnitude.

the relationship between the output torque of the exoskeleton and the motor current is straightforward. Such a relationship can be modeled as Equation (4), where $T$ is the output torque, $I$ is the current of the motor, $k$ is the torque constant, and $T_f$ is the friction torque. By means of model identification, the output torque can be precisely inferred from the prior information, and the accurate torque estimation allows the direct torque control of our exoskeleton. One of the substantial benefits of the control method is that the exoskeleton can be more lightweight and compact because of the avoidance of torque sensors.

$$T = \begin{cases} kI - T_f, & I > T_f/k \\ kI + T_f, & I < -T_f/k \\ 0, & -T_f/k \leq I \leq T_f/k \end{cases} \quad (4)$$

Fig. 13 demonstrates the parameter estimation results of model (4). The test was conducted on a testbed with high-precision force sensor (LRM200, Futek, Inc.) to measure the output torque. Through linear regression, the torque constant $k$ and friction torque $T_f$ are estimated as 0.62 Nm/A and 0.5 Nm, respectively. The R2 coefficient is 96.14%. Based on this model, our actuator direct torque control can track the 10 Hz and 15 Nm sine reference trajectory in the torque control test with 0.88 Nm root mean square error, as shown in Fig. 14. Because the frequency range of human motion is generally lower than 10 Hz, this result indicates that the torque control bandwidth of our exoskeleton is sufficient for the human assistance.

### B. Mechanical Transparency Evaluation in Passive Mode

To investigate the backdrivability of the exoskeleton, the resistive torque of the exoskeleton in passive mode was evaluated and the result is shown in Fig. 15. The passive mode torque is within ±2 Nm and it reveals that our low transmission ratio design is highly-backdrivable and transparent, ensuring a safe interaction between human and the robot. Compared to traditional high transmission ratio actuation systems that cannot be back-driven by the user [51], [52] this solution presents a safer and more viable device design.

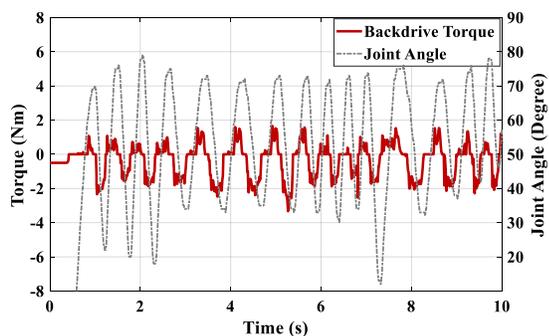

Fig. 15. The no-load impedance test for the transparency evaluation. The maximum amplitude of torque is 3.33 Nm and the root mean square of the torque is 1.03 Nm.

### C. Torque Tracking Evaluation with Three Human Subjects

The assistive control for the stance phase was evaluated in three healthy subjects. As shown in the Fig. 16, the exoskeleton assisted the knee extension from flat foot phase to heel-off phase. When the angular velocity of the ankle joint starts to become stable around zero (A), the gait cycle detection algorithm detects the start of stance phase, and the assistive torque profile is triggered (C). When the ankle angular velocity starts to change noticeably, the algorithm detects the end of the stance phase (C). Then the assistive torque ends at d. The bottom of Fig.16 demonstrates the torque tracking performance of the exoskeleton. The root mean the tracking square error on all the subjects is 0.31 Nm.

## V. DISCUSSION AND CONCLUSIONS

In this paper, we presented a body-worn knee exoskeleton design to overcome discomfort issues. A new wearable structure design is analyzed, optimized and compared with conventional methods. The transmission design ensures reduced joint misalignment and low mechanical impedance. Consistent with our design performance, flexible exoskeletons developed by Samsung [26] also exhibit minimal interference with human motion and low impedance. Comparing with the prior art in knee exoskeletons [53]–[56], this design focuses on human-centered design solutions to develop comfortable and safe personal mobility assistants. The assistive control scheme is force control based with salient benefits for human-robot interaction comparing with exoskeletons under position control. The simulation and experimental results demonstrate the feasibility and effectiveness of our exoskeleton design with reduced joint misalignment, small resistive torque in unpowered mode, and consistent torque tracking performance. Considering the feasibility results, it is planned to further reduce the weight of the system with lightweight structural materials, reduce the thickness of 2-stage transmission and increase the torque output. Since the system is lightweight and creates consistent torque assistance without disrupting the natural gait, we expect to improve control strategy and conduct human studies to investigate its biomechanical effects.

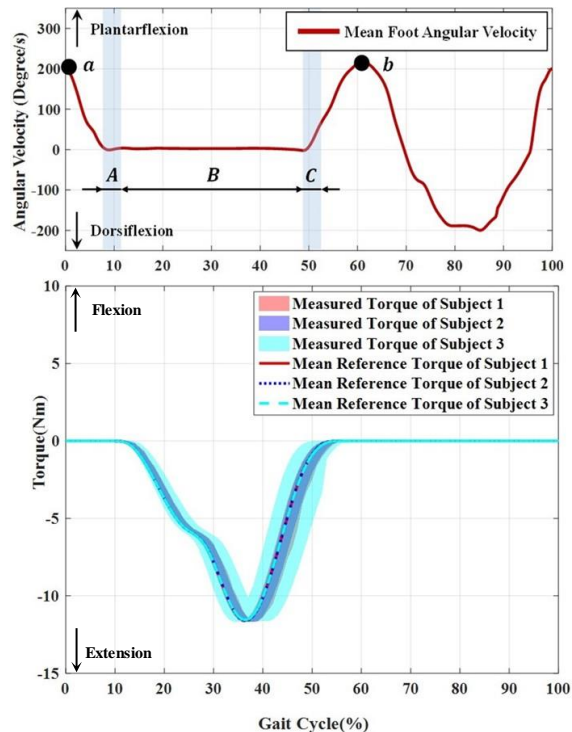

Fig. 16. (Top) Foot angular velocity-based gait detection. A indicates the detection window for stance phase. B indicates stance phase. C indicates the detection window for stance phase end point. When the algorithm detects the foot flat phase, the assistive torque profile is triggered. (Bottom) the shaded area illustrates the torque tracking performance in 100 strides of three different human subjects. The assistive torque ends at the heel-off event. The root mean square error of torque tracking on three human subjects is 0.31 Nm.